\documentclass[runningheads]{llncs}
\usepackage{graphicx}
\usepackage{booktabs} 
\usepackage{tabularx}
\usepackage{arydshln} 
\usepackage{enumerate}
\usepackage{amssymb}
\usepackage{amsmath}
\usepackage[dvipsnames]{xcolor}
\usepackage[binary-units=true]{siunitx}
\usepackage[hidelinks]{hyperref}
\usepackage{algorithm}
\usepackage{algpseudocode}
\usepackage[bottom]{footmisc}  

\begin{document}
\title{Inferring, Predicting, and Denoising Causal Wave Dynamics\vspace{-0.2cm}}

%
%
\author{
	Matthias Karlbauer\inst{1}\orcidID{0000-0002-4509-7921}\and\\
	Sebastian Otte\inst{1}\orcidID{0000-0002-0305-0463}\and\\
	Hendrik P.A. Lensch\inst{2}\and\\
	Thomas Scholten\inst{3}\and\\
	Volker Wulfmeyer\inst{4}\and\\
	Martin V. Butz\inst{1}\orcidID{0000-0002-8120-8537}
}
\authorrunning{M. Karlbauer et al.}
%
\institute{
	University of T\"ubingen -- Neuro-Cognitive Modeling Group,\\
	Sand 14, 72076 T\"ubingen, Germany, \email{martin.butz@uni-tuebingen.de}\thanks{This work received funding from the DFG Cluster of Excellence ``Machine Learning: New Perspectives for Science'', EXC 2064/1, project number 390727645. Moreover, we thank the International Max Planck Research School for Intelligent Systems (IMPRS-IS) for supporting Matthias Karlbauer.} 
	\and
	University of T\"ubingen -- Computer Graphics,\\
	Maria-von-Linden-Stra\ss e 6, 72076 T\"ubingen, Germany 
	\and
	University of T\"ubingen -- Soil Science and Geomorphology,\\
	R\"umelinstra\ss e 19-23, 72070 T\"ubingen, Germany 
	\and
	University of Hohenheim -- Institute for Physics and Meteorology,\\
	Garbenstra\ss e 30, 70599 Stuttgart, Germany
	\vspace{-0.2cm}
}
\maketitle              
%

\begin{abstract}
	The novel DISTributed Artificial neural Network Architecture (DISTANA) is a generative, recurrent graph convolution neural network. 
    It implements a grid or mesh of locally parameterizable laterally connected network modules. DISTANA is specifically designed to identify the causality behind spatially distributed, non-linear dynamical processes.
    We show that DISTANA is very well-suited to denoise data streams, given that re-occurring patterns are observed, significantly outperforming alternative approaches, such as temporal convolution networks and ConvLSTMs, on a complex spatial wave propagation benchmark.
	It produces stable and accurate closed-loop predictions even over hundreds of time steps.
	Moreover, it is able to effectively filter noise---an ability that can be improved further by applying denoising autoencoder principles or by actively tuning latent neural state activities retrospectively. 
	Results confirm that DISTANA is ready to model real-world spatio-temporal dynamics such as brain imaging, supply networks, water flow, or soil and weather data patterns.
\end{abstract}
%

\section{Inroduction}

Although sufficiently complex artificial neural networks (ANNs) are considered as universal function approximators, the past has shown that major advances in the field of artificial intelligence were frequently grounded in specific ANN structures that were explicitly designed to solve a particular task, such as long short-term memories (LSTMs) for time series prediction, convolutional neural networks (CNNs) for image processing, or autoencoders (AEs) for data compression.
This illustrates that---although theoretically any ANN can solve any desired task---a network model benefits considerably from being reasonably restricted in a way that constrains (and thus limits) the possibilities of approximating a desired process.

The process that is to be modeled here, is a two-dimensional circular wave expanding from a point source---a spatio-temporal learning problem that has been shown to be challenging for known ANN architectures. 
In \cite{karlbauer2019distributed}, however, a novel distributed graph neural network (GNN) architecture was introduced, named DISTANA, that is specifically designed to learn these kinds of data appropriately.
It was demonstrated that this architecture can accurately predict spatio-temporal processes over hundreds of closed loop steps into the future, while assuming that the underlying data has a dynamic graph-like structure.

GNNs raise new challenges as traditional convolution techniques are not directly applicable, but they also offer new possibilities as they explicitly facilitate the processing of irregularly distributed data patterns \cite{shuman2013emerging}.
Accordingly, various promising GNNs have been proposed recently \cite{scarselli2008graph,battaglia2018relational}, which explicitly encode graph nodes (vertices) and node connections (edges), making them applicable to a wide range of problems.
\cite{wu2019comprehensive} offers a general survey on GNNs and distinguishes between convolutional GNNs, graph autoencoders, and spatio-temporal GNNs (called spatial-temporal in \cite{wu2019comprehensive}, STGNNs).
DISTANA is a kind of STGNN and shares architectural principles with \cite{jain2016structural}.

The contribution of this paper is the application of DISTANA in order to filter noise from an underlying dynamic, spatially distributed data stream without latency.
To accomplish this, two techniques are investigated: (a) training models directly on noisy data and (b) applying active tuning \cite{active_tuning_patent_2019}, a retrospective, gradient based inference technique.
Results are compared to convolutional LSTMs (ConvLSTMs) \cite{xingjian2015convolutional} and temporal convolution networks (TCNs) \cite{bai2018empirical}. 
While TCNs have been proposed as being more suitable than recurrent neural networks (RNNs) in modeling time series data, ConvLSTM constitutes a convolution-gating LSTM architecture, which is able to perform systematic video predictions, including the dynamics in sequential satellite images of clouds.
DISTANA models the wave benchmark more accurately in mid to low noise conditions and generalizes to larger extent.
Furthermore, active tuning filters noise from distributed sensor meshes highly effectively, outperforming standard denoising approaches.

%
%

\section{Methods}

\begin{figure}[t]
	\centering
	\includegraphics[width=\textwidth]{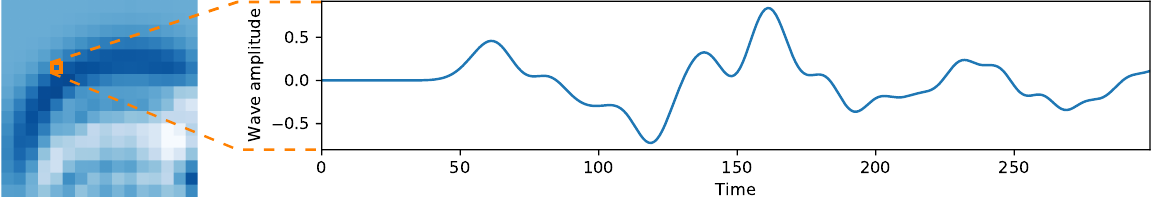}
    \vspace{-0.2cm}
	\caption{Left: $16\times16$ data grid example, showing a circular wave around time step 55 propagating from bottom right to top left and being reflected at the borders. Right: the wave amplitude dynamics over time for one pixel in the 2D wave field.}
	\label{methods:fig:data_description}
\end{figure}

In this paper, we focus on modeling and, in particular, denoising a two-dimensional circular wave benchmark, which expands in space over time while being reflected at borders, as illustrated in  \autoref{methods:fig:data_description}.
Wave dynamics are generated by a set of local differential equations as specified in \cite{karlbauer2019distributed}.

For training, the standard MSE metric is used, whereas the dynamic time warping (DTW) distance \cite{salvador2007toward} is used to evaluate and test the models.
The DTW distance compares two sequences by finding the smallest number of transition operations to transfer one sequence into the other.
When noise is added to the data (to be filtered away by the model), it is reported in terms of its signal to noise ratio (SNR), which is computed as $\text{SNR} = P_\text{signal}/P_\text{noise}$, where the power of the signal $P_\text{signal}$ and of the noise $P_\text{noise}$ are calculated as root mean square amplitudes (RMSA), using
$\text{RMSA} = \sqrt{\frac{1}{T\cdot I\cdot J}\sum_{t}\sum_{i}\sum_{j}s_{tij}^2}$,
where $t$ denotes the time, $i$ and $j$ the position, and $s$ the signal value.

Three different types of artificial neural network models were implemented.
For each of these three types, two versions were tested: a small variant of roughly $300$ parameters and a large variant with roughly $3\,000$ parameters. Numerous model complexities were compared.
We report the best results throughout.

\subsection{Convolutional LSTM (ConvLSTM)}
Both ConvLSTM variants consist of two ConvLSTM layers \cite{xingjian2015convolutional}.
The simple version, ConvLSTM2 with 324 parameters, projects the $16\times16\times1$ input via the first layer on two feature maps (resulting in dimensionality $16\times16\times2$) and subsequently via the second layer back to one output feature map.
All kernels have a filter size of $k = 3$, use zero-padding and come with a stride of one.
The complex version, ConvLSTM8 with $2\,916$ parameters, projects the input to eight feature maps and subsequently back to one output map.
Code was taken and adapted from \footnote{\url{https://github.com/ndrplz/ConvLSTM_pytorch}}.

\subsection{Temporal Convolution Network (TCN)}
The two TCN variants have the same principal structure: an input layer projects to either two (TCN121, 320 parameters) or to nine (TCN191, $2\,826$ parameters) feature maps, which project their values back to one output value.
A kernel filter size of $k = 3$ is used for the two spatial dimensions in combination with the standard dilation rate of $d = 1, 2, 4$ for the temporal dimension, resulting in a temporal horizon of 14 time steps (cf. \cite{bai2018empirical}).
Deeper networks with larger temporal horizon did not improve performance.
Code was taken and adapted from \cite{bai2018empirical}.

\subsection{Our spatio-temporal graph neural network (DISTANA)}

\begin{figure}[b]
	\centering
	\includegraphics[height=3.65cm]{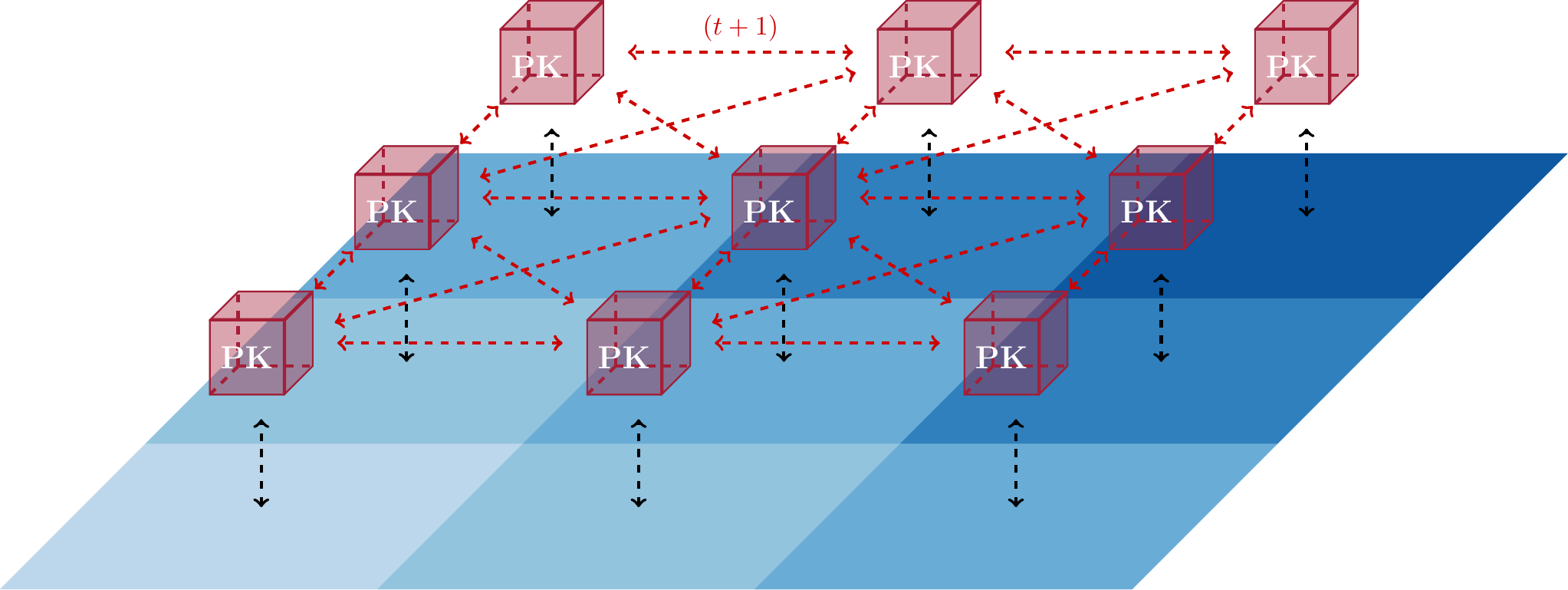}\hfill
	\includegraphics[height=3.65cm]{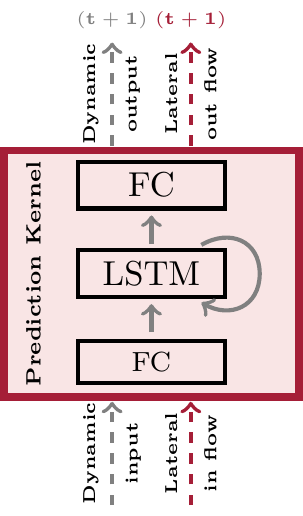}
	\caption{Left: $3\times3$ sensor mesh grid visualizing a 2D wave propagating from top right to bottom left and showing the connection scheme of Prediction Kernels (PKs) that model the local dynamical process while communicating laterally. Right: exemplary prediction kernel, receiving and predicting dynamic and lateral information flow.}
	\label{methods:fig:distana_architecture}
\end{figure}

DISTANA \cite{karlbauer2019distributed} assumes that the same dynamic principles apply at every spatial position.
Accordingly, it models the data at each position in a regular sensor mesh with the identical network module.
This module, which we call Prediction Kernel (PK), consists of a 4-neurons, $\tanh$-activated fully connected layer (compressing one dynamic input value and eight lateral input values), followed by an LSTM layer with either four (DISTANA4, 200 parameters) or 24 (DISTANA24, $2\,940$ parameters) LSTM cells, followed by a 9-neurons $\tanh$-activated fully connected layer (producing one dynamic output prediction and eight lateral outputs).
Each PK predicts the dynamics of its assigned pixel (or sensor mesh point) in the next time step, given its previous hidden state, the predicted or measured sensor data, and the lateral information coming from the PKs in the direct eight-neighborhood (see~\autoref{methods:fig:distana_architecture}).
Standard LSTM cells are used \cite{hochreiter1997long}. 
However, no bias neurons are used to prevent spontaneous cell activation without external input. 


More concretely, the DISTANA version considered here consist of a prediction kernel (PK) network, which is made up of the weight matrices $\mathbf{W}^{DL}_\text{pre}$, representing a dynamic- and lateral input preprocessing layer, the weight matrices of a regular LSTM layer (see \cite{hochreiter1997long}), and $\mathbf{W}^{DL}_\text{post}$, defining a dynamic- and lateral output postprocessing layer.
DISTANA implements $k\in\mathbb{N}$ PK instances ${\{p_1, p_2, \dots, p_k\}}$, where each one has a list of eight spatial neighbors $\mathbf{n}_i\in\mathbb{N}^8$. 
Each PK instance $p_i$ receives dynamic input $\mathbf{d}_i\in\mathbb{R}^d$ and lateral input $\mathbf{l}_i\in\mathbb{R}^l$ from the eight neighboring PK instances, with $d, l\in\mathbb{N}$ being the dimensionality of dynamic- and lateral inputs, respectively.
The LSTM layer in each PK instance contains $m\in\mathbb{N}$ cells with hidden- and cell states $\mathbf{h}_i$ and $\mathbf{c}_i\in\mathbb{R}^m$.

To perform the forward pass of PK instance $p_i$ at time step $t$, the corresponding dynamic- and lateral inputs $\mathbf{d}_i^{t-1}, \mathbf{l}_i^{t-1}$, with $\mathbf{l}_i^{t-1}=\{\mathbf{l}_i|i\in\mathbf{n}_i\}$, along with the according LSTM hidden- and cell states $\mathbf{h}_i^{t-1}, \mathbf{c}_i^{t-1}$, are fed into the PK network to realize the following computations:
\begin{align}
	\mathbf{dl}_\text{pre} & = {\tanh}{\left(\mathbf{W}^{DL}_{\text{pre}} (\mathbf{d}^{t-1}_i\circ \mathbf{l}^{t-1}_i)\right)}\label{methods:eq:dl_pre_computation}\\
	\mathbf{c}^t_i, \mathbf{h}^t_i & = LSTM(\mathbf{dl}_\text{pre}, \mathbf{c}^{t-1}_i, \mathbf{h}^{t-1}_i)\\
	\mathbf{dl}_\text{post} & = {\tanh}{\left(\mathbf{W}^{DL}_\text{post} \mathbf{h}^t_i\right)}\label{methods:eq:dl_post_computation}\\
	[\mathbf{d}^t_i|\mathbf{l}^t_i] & = \mathbf{dl}_\text{post},
\end{align}
where vector concatenations are denoted by the $\circ$ operator.
As depicted in \autoref{methods:eq:dl_pre_computation} and \autoref{methods:eq:dl_post_computation}, the lateral input and output is processed by a fully connected nonlinear layer, leading to a potentially different treatment of the lateral information coming from each direction.
Although transitions between neighboring PKs should, in theory, be direction invariant, we give the model the freedom to develop unique transition weights. 
Experiments with shared perpendicular and diagonal weights did not improve performance (not shown).

\subsection{Active Tuning}\label{methods:subsec:active_tuning}
Active tuning (AT) is a technique that allows to use an already trained network for signal denoising, input reconstruction, or prediction stabilization, even if the training did not cover these tasks \cite{active_tuning_patent_2019}.
In this work, AT is used to filter noise by inducing reasonable activity into the network, meaning that a dynamic input to the PKs is inferred such that they reflect the current wave pattern of the dynamic process (see Algorithm~\ref{alg:at}).

\begin{algorithm}[b]
\caption{Active Tuning procedure}\label{alg:at}
\begin{algorithmic}[1]
\State Initialize zero or random input vector $\mathbf{x}$.
\State Forward $\mathbf{x}$ through the network to obtain the network output vector $\mathbf{y}$.
\State Compute and apply gradients on $\mathbf{x}$ by comparing the network output vector $\mathbf{y}$ with the noisy target vector $\mathbf{t}$.
\State Repeat (2) and (3) until convergence or for $c$ optimization cycles.
\end{algorithmic}
\end{algorithm}

Technically, AT differs from teacher forcing (which traditionally is used for activity induction) in that AT prevents the network from receiving data directly.
Instead, the input is inferred via prediction error induced temporal gradients, while comparing the output of the network to some target signal.
Consequently, the gradient-based inferred input can only take values which the network itself can generate and thus also interpret correctly.
If the target sequence is noisy but the network has never been trained to generate noise, the inferred input will only consist of plausible, i.e. known signal components, which effectively filters implausible or unknown characteristics, such as noise, from the target.

The signal filtering requires careful tuning of basically two parameters: $H$ as the history length, which indicates how many time steps into the past the input vector $\mathbf{x}$ will be projected and optimized, and $\eta$ as the learning rate, which weighs the update of $\mathbf{x}$ based on the obtained gradients.
A third potential parameter is the number of optimization cycles $c$, that is, how often the optimization procedure is repeated.
We set $c=30$ in this work.

%

\section{Experiments and Results}

All models were trained for 200 epochs with 100 training sequences of length 40 each.
Every model was trained ten times with different random seeds to obtain mean and standard deviation scores.
A learning rate of $0.001$ was used in combination with the ADAM optimizer \cite{kingma2014adam}, minimizing the MSE between network output and target.
Different noise levels were added to the data and the signal to noise ratios (SNRs) $[\SI{0.25}{},\allowbreak \SI{0.5}{},\allowbreak \SI{1}{},\allowbreak \SI{2}{},\allowbreak \SI{4}{},\allowbreak \SI{10}{},\allowbreak \SI{100}{},\allowbreak \SI{e+3}{},\allowbreak \SI{e+5}{}]$ were evaluated, where \SI{0.25}{} is a particularly challenging case (the signal has only a quarter of the power of the noise) and \SI{e+5}{} is considered as the noise-free case. 

Model evaluations are based on the dynamic time warping (DTW) distance 
on 20 test sequences consisting of a spatial size of $16\times16$ pixels and 150 time steps.
Either 30 teacher forcing steps or 30 active tuning steps are initially performed (to induce activity into the network), followed by 120 closed loop steps (to evaluate the actual model accuracy at continuing the spatio-temporal wave dynamics).
The DTW distance metric is exclusively calculated over the 120 closed loop steps to measure the stability of each model when running detached from external data input.
Note that all models were only trained on 40 time steps and, in consequence, are required to properly generalize over 150 time steps in order to reach a small test error.
More noise-free experiments about comparing DISTANA to many more baseline approaches can be found in \cite{karlbauer2019distributed}.

\subsection{Training on noisy data -- ConvLSTM vs TCN vs DISTANA}\label{experiments_and_results:subsec:training_on_noisy_data}
To quantify the different network model's denoising capabilities 
when being trained explicitly on noisy data, ten models of each variant were trained on the different SNRs specified above.
We expect that an increasing SNR results in lower error rates, since in low noise cases (high SNR) a model does not have to additionally filter noise disturbances to encounter the actual signal.

\begin{figure}[t]
	\centering
	\includegraphics[width=0.49\textwidth]{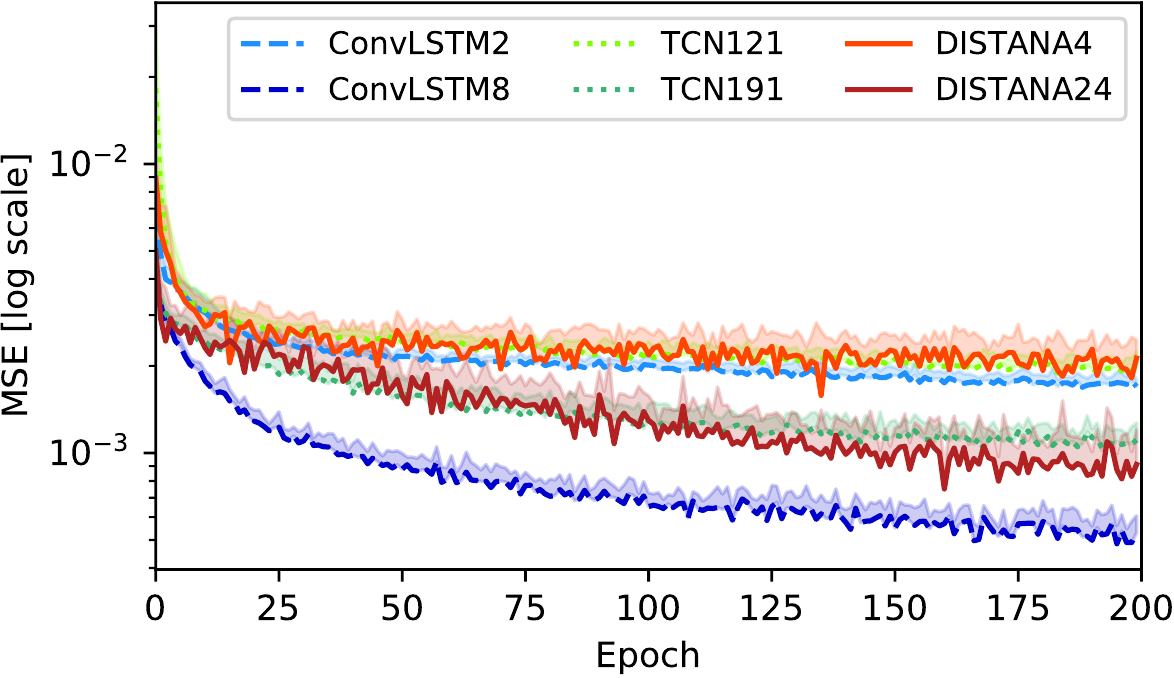}
	\hfill
	\includegraphics[width=0.49\textwidth]{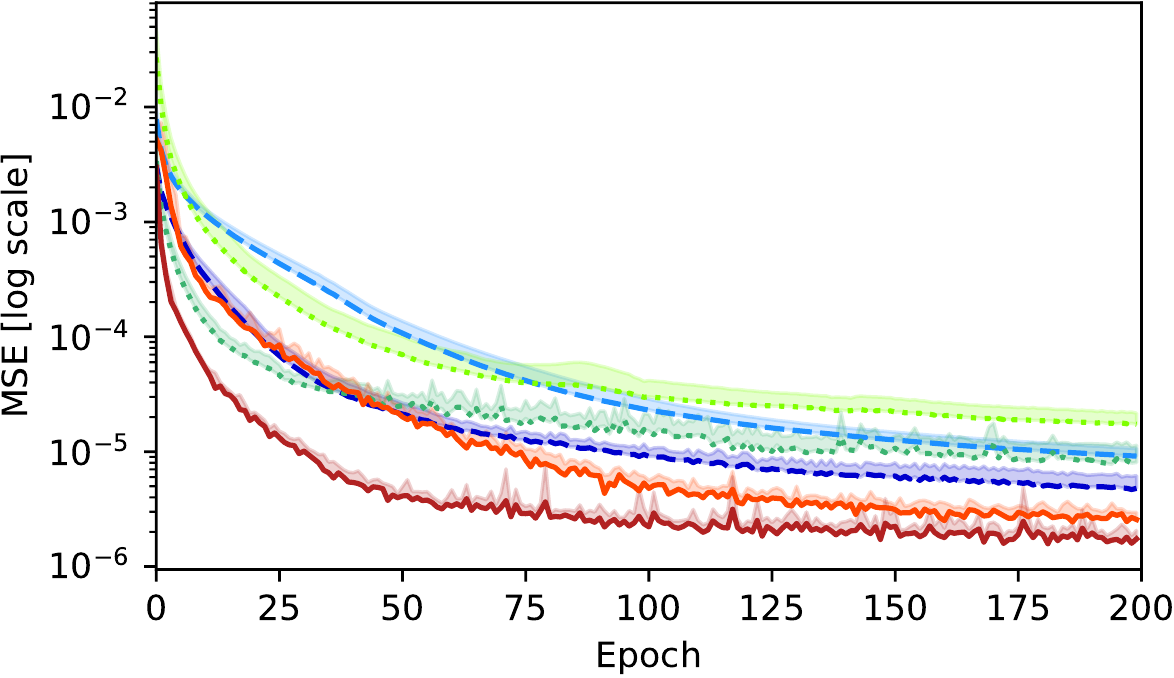}
    \vspace{-0.2cm}
    \caption{Training convergence on different signal to noise ratios (left: \SI{0.25}{}, right: \SI{e+5}{}) for two versions of each of the compared network architectures ConvLSTM, TCN and DISTANA. Note the logarithmic y-axis and different y-axis scales.}
	\label{experiments_and_results:fig:training_convergence}
\end{figure}

\subsubsection{Training convergence}
Two different convergence plots, showing the decreasing MSE error over time along with standard deviations for all models, are provided in \autoref{experiments_and_results:fig:training_convergence}.
While the left convergence plot, depicting training on the challenging \SI{0.25}{SNR} condition, clearly favorites the large ConvLSTM8 model over all others, both DISTANA variants outperform the ConvLSTM and TCN models on the noise-free condition, that is \SI{e+5}{SNR}, as depicted in the right convergence plot.

\begin{figure}[!b]
	\centering
	\includegraphics[width=0.49\textwidth]{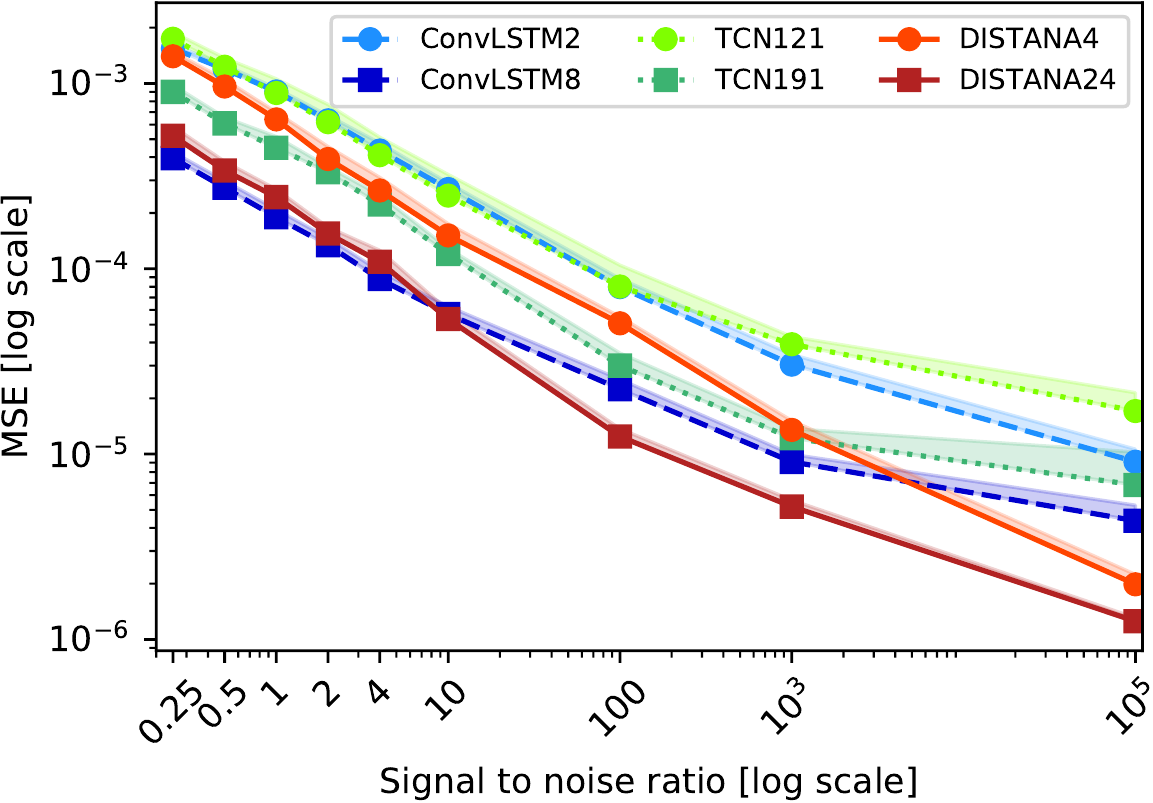}
	\hfill
	\includegraphics[width=0.49\textwidth]{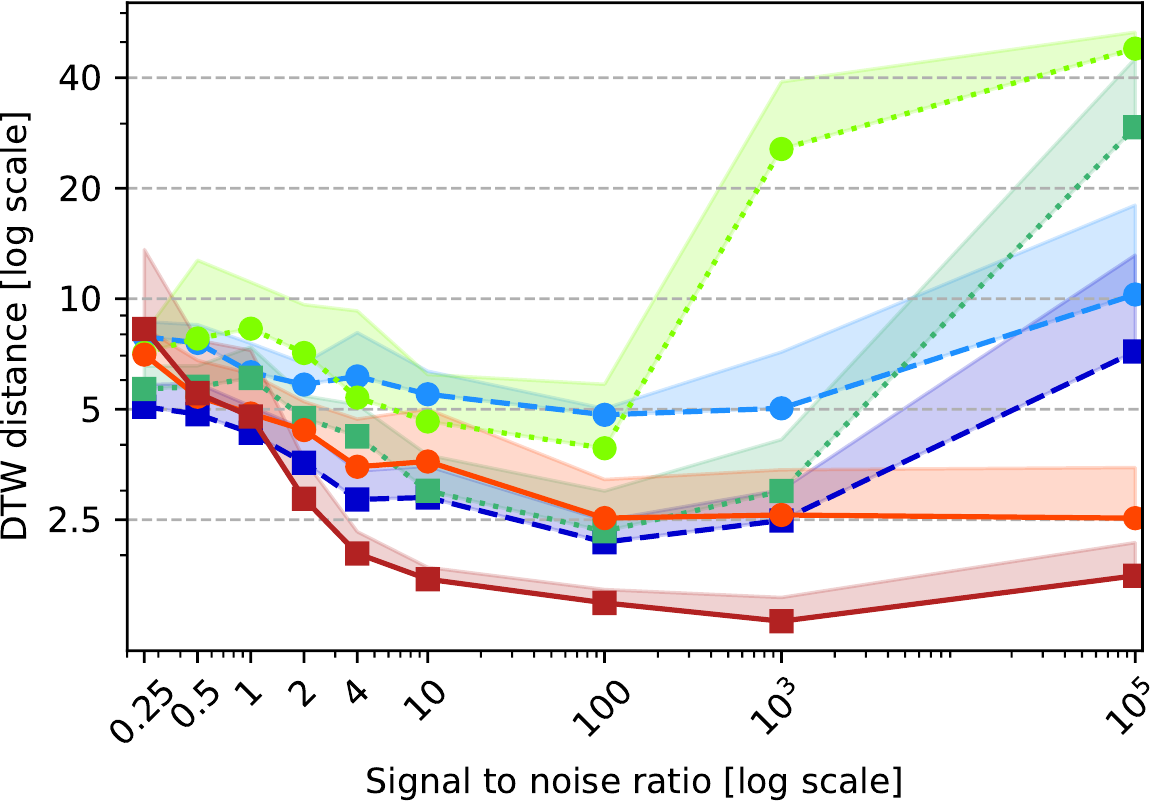}	
    \vspace{-0.2cm}
    \caption{Left: final training MSE scores over different SNRs for two versions---300 parameters (circular markers, bright color), $3\,000$ parameters (square markers, dark color)---of each of the three compared architectures ConvLSTM (blues, dashed), TCN (greens, dotted), DISTANA (reds, solid). Note the logarithmic scales on both x- and y-axis. Right: test accuracies, based on dynamic time warping (DTW) distance.}
	\label{experiments_and_results:fig:mse_and_dtw_accuracy_comparison}
\end{figure}

A further analysis reasonably shows the expected trend that large SNRs (small noise levels) lead to lower and thus better training errors (see \autoref{experiments_and_results:fig:mse_and_dtw_accuracy_comparison}, left).
The same plot also reveals how much the training approximation accuracy of a certain model depends on the SNR, confirming the previous finding that ConvLSTM seems to be slightly better than other models for a low SNR (large noise) while DISTANA considerably reaches superior performance on low-noise conditions (high SNR).

\subsubsection{DTW-based model evaluation}
A model evaluation based on the DTW distance is visualized in the right plot of \autoref{experiments_and_results:fig:mse_and_dtw_accuracy_comparison}.
These results corroborate the findings from the training analyses and emphasize the GNN's superiority over the other two approaches even more clearly.
With an order of magnitude less parameters, DISTANA4 reaches the same level of performance as the large ConvLSTM and TCN architectures, while, with an equal number of parameters, the GNN model produces significantly better results compared to ConvLSTM and TCN (note the logarithmic y-axis scale, making the performance differences even clearer).
Additional video material\footnote{\url{https://youtu.be/j8xJnoo1wOo}} shows that none of the models reaches satisfying results on the lowest SNR condition when trained directly on noisy data, which is noticeable in either a quick fading to zero or a chaotic wave activity; an issue that we approach in the following section.

\subsection{Active Tuning}\label{experiments_and_results:subsec:active_tuning}
As mentioned beforehand (see \autoref{methods:subsec:active_tuning}), active tuning (AT) can be used, for example, to infer network inputs via gradients.
Therefore, AT can replace teacher forcing (TF), where data are fed directly into the network to induce adequate activity into the model and thus to let it run in closed loop to produce future predictions.

\begin{table}[!b]
	\centering
	\caption{Dynamic time warping distances for models that are initialized with teacher forcing (TF) or active tuning (AT) and that have been trained and tested on different signal to noise ratios (SNR).
	Varied AT parameters $\eta$ and $H$ are reported for each model. 
	Entries above the dashed line in each Train SNR block correspond to models that were tested on larger noise than they were trained on.
	Superior performances are emphasized in blue.}
	\label{experiments_and_results:tab:active_tuning}
	\vspace{-0.2cm}
	\begin{small}
{
\newcommand{\tabemph}[1]{\color{Blue}#1}
\begin{tabular*}{\textwidth}{
        l
		S[table-format=2.2(4), separate-uncertainty=true]@{\extracolsep{\fill}}
		S[table-format=2.2(3), separate-uncertainty=true]@{\extracolsep{\fill}}
		S[table-format=1.3]@{\extracolsep{\fill}}
		S[table-format=2]@{\extracolsep{\fill}}
		S[table-format=2.2(4), separate-uncertainty=true]@{\extracolsep{\fill}}
		S[table-format=2.2(3), separate-uncertainty=true]@{\extracolsep{\fill}}
		S[table-format=1.3]@{\extracolsep{\fill}}
		S[table-format=2]@{\extracolsep{\fill}}
	}
	\toprule
	Test SNR & TF & AT & $\eta$ & $H$ & TF & AT & $\eta$ & $H$\\
	\midrule
	& \multicolumn{4}{l}{Train SNR: $0.25$} & \multicolumn{4}{l}{Train SNR: $4$}\\
	\cmidrule{2-5}\cmidrule{6-9}
	$0.25$ & 8.02(101) & \tabemph{7.11(098)} & 0.400 & 1 & 63.71(236) & \tabemph{5.78(143)} & 0.020 & 25 \\\cdashline{6-9}
	$4.00$ & 8.25(101) & \tabemph{7.78(103)} & 0.350 & 2 & 5.50(137) & \tabemph{5.26(135)} & 0.100 & 25 \\
	\SI{100}{} & 8.25(102) & \tabemph{7.94(115)} & 0.500 & 5 & \tabemph{4.95(128)} & 5.08(133) & 0.100 & 25 \\
	\SI{e+5}{} & 8.25(102) & \tabemph{7.94(115)} & 0.500 & 5 & \tabemph{4.91(129)} & 5.01(121) & 0.500 & 1 \\
	\midrule
	& \multicolumn{4}{l}{Train SNR: \SI{e+2}{}} & \multicolumn{4}{l}{Train SNR: \SI{e+5}{}}\\
	\cmidrule{2-5}\cmidrule{6-9}
	$0.25$ & 38.08(1219) & \tabemph{3.58(060)} & 0.010 & 25 & 23.86(091) & \tabemph{8.26(049)} & 0.002 & 25 \\
	$4.00$ & 3.49(039) & \tabemph{3.00(060)} & 0.070 & 10 & 15.31(127) & \tabemph{4.96(076)} & 0.004 & 25 \\\cdashline{2-5}
	\SI{100}{} & \tabemph{1.93(017)} & 2.09(026) & 0.400 & 4 & 6.06(057) & \tabemph{4.52(090)} & 0.005 & 25 \\\cdashline{6-9}
	\SI{e+5}{} & \tabemph{1.76(018)} & 1.79(019) & 0.300 & 1 & \tabemph{4.14(047)} & 4.25(048) & 0.300 & 2 \\
	\bottomrule
\end{tabular*}
}

	\end{small}
\end{table}

Here, we compare DISTANA4's closed loop performance when initial activities are induced via TF or AT. DISTANA4 was chosen due to its small number of parameters while reaching comparably good performance; however, the results in principle can be generalized to any desired recurrent neural network model. TF and AT are compared in three conditions: (a) the SNR during training is larger than the SNR during testing, that is $\text{SNR}_\text{train} > \text{SNR}_\text{test}$, (b) $\text{SNR}_\text{train} = \text{SNR}_\text{test}$, and (c) $\text{SNR}_\text{train} < \text{SNR}_\text{test}$. The four SNRs $[\SI{0.25}{},\allowbreak \SI{4}{},\allowbreak \SI{100}{},\allowbreak \SI{e+5}{}]$ are compared (see \autoref{experiments_and_results:tab:active_tuning}).

\subsubsection{Stability and applicability gains through AT}
As indicated by the blue-colored entries in \autoref{experiments_and_results:tab:active_tuning}, AT outperforms TF in 11/16 cases. In most cases, however, the techniques do not differ significantly.
Yet, in condition (c), DISTANA4 systematically benefits from being initialized through AT, that is if $\text{SNR}_\text{train} < \text{SNR}_\text{test}$.
Referring to \autoref{experiments_and_results:tab:active_tuning}, these cases are above the dashed horizontal lines within each Train SNR block.
In consequence, AT makes a model applicable to larger noise than it was trained on.
Besides, using AT instead of TF never decreases the model performance significantly.

\begin{figure}[t]
	\centering
	\includegraphics[width=\columnwidth]{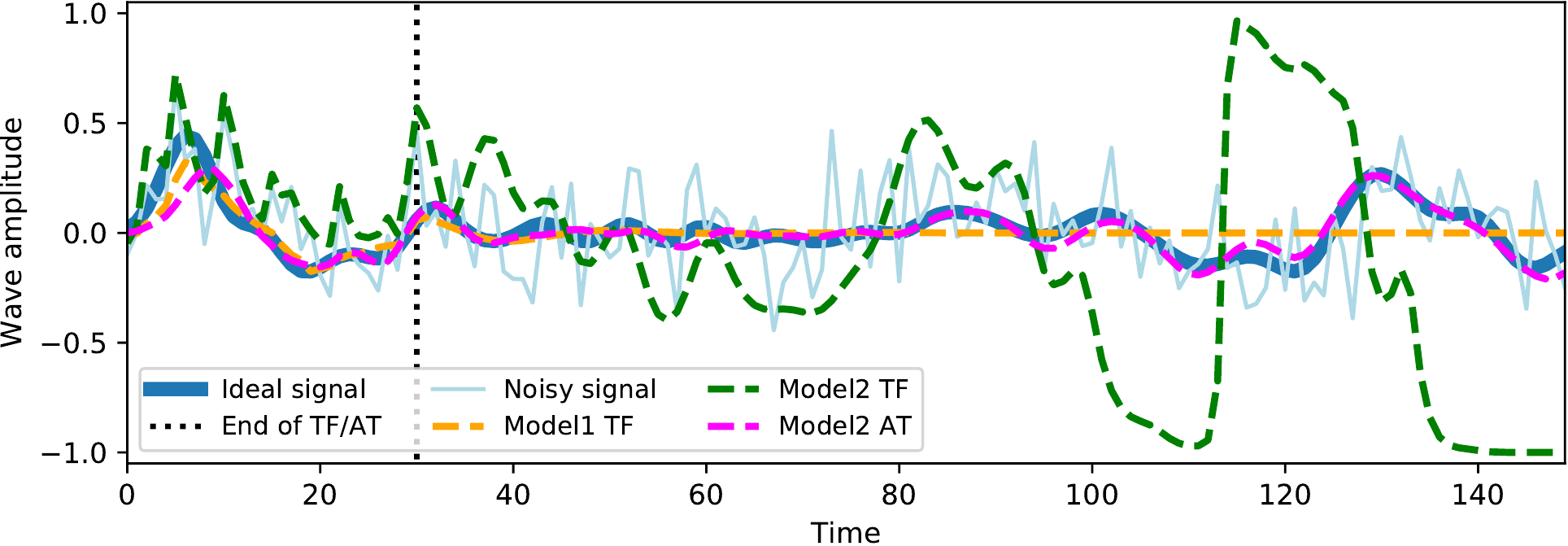}
	\vspace{-0.2cm}
	\caption{Denoising a signal (light blue) with \SI{0.25}{SNR} using either teacher forcing (TF) or active tuning (AT) to induce reasonable activity in the first 30 time steps. Model1 is trained on \SI{0.25}{SNR} and tested with TF (orange, dashed). Model2 is trained on \SI{100}{SNR} and tested with TF (dark green, dashed) and with AT (magenta, dashed). Ideal signal displayed in dark blue; all models are DISTANA4 architectures.}
	\label{experiments_and_results:fig:active_tuning_activity}
\end{figure}

\begin{figure}[!b]
	\centering
	\includegraphics[width=0.193\columnwidth]{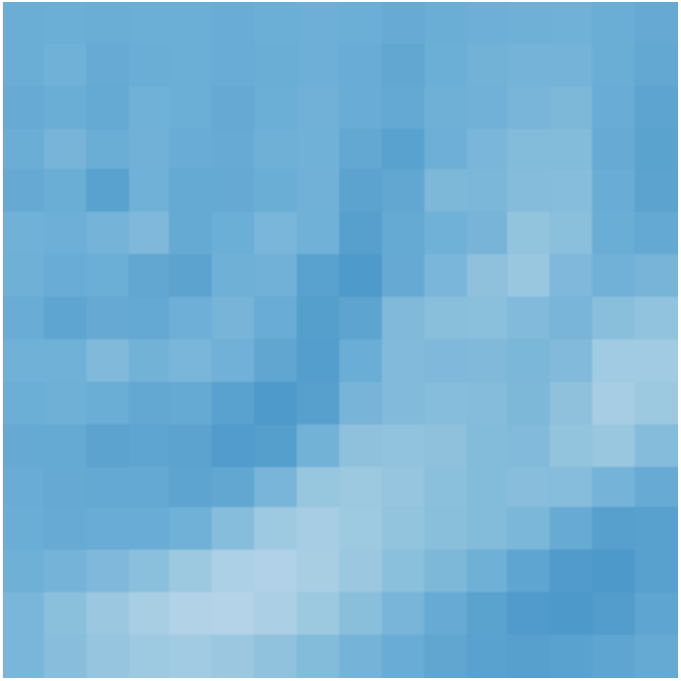}
	\includegraphics[width=0.193\columnwidth]{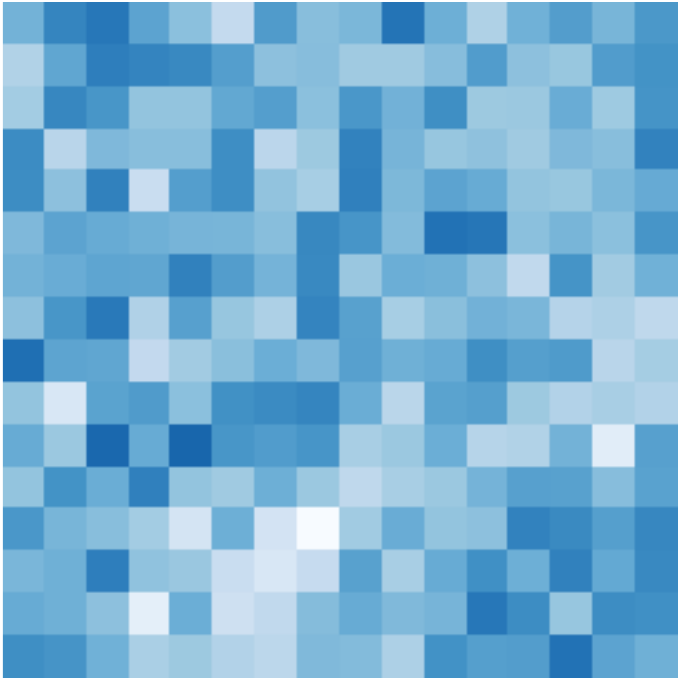}
	\includegraphics[width=0.193\columnwidth]{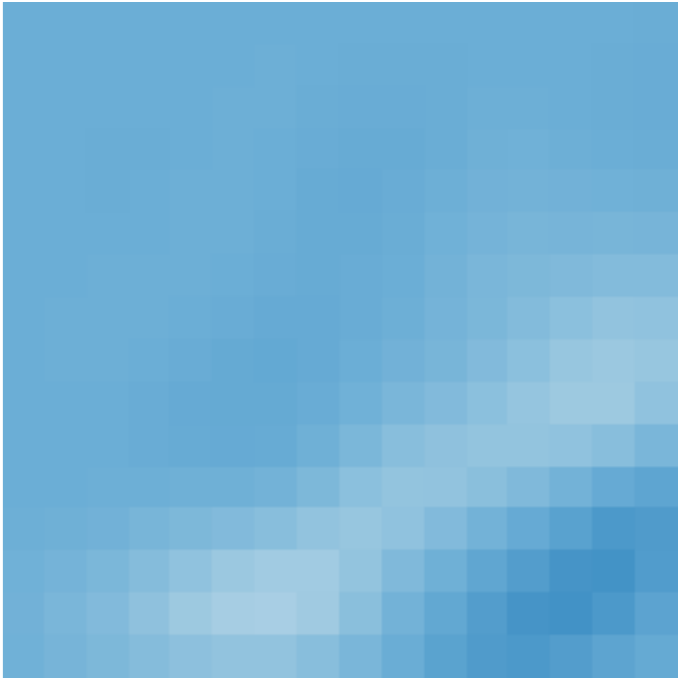}
	\includegraphics[width=0.193\columnwidth]{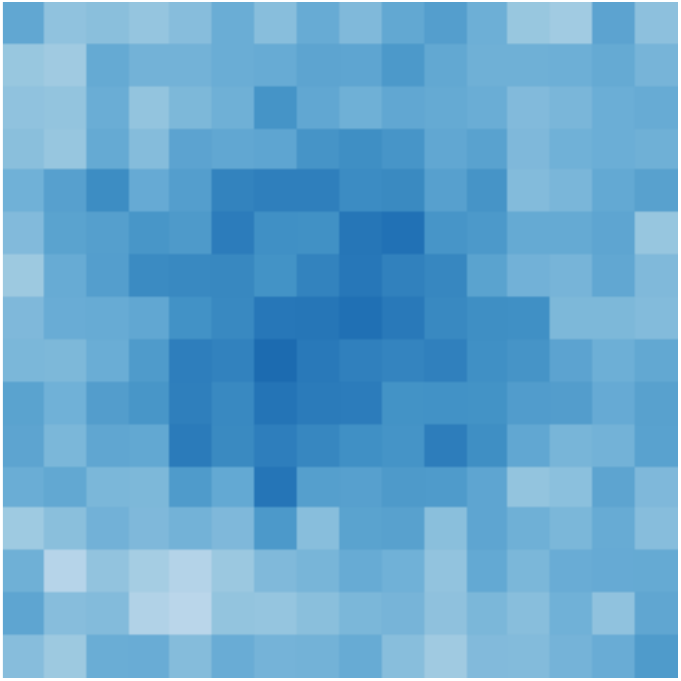}
	\includegraphics[width=0.193\columnwidth]{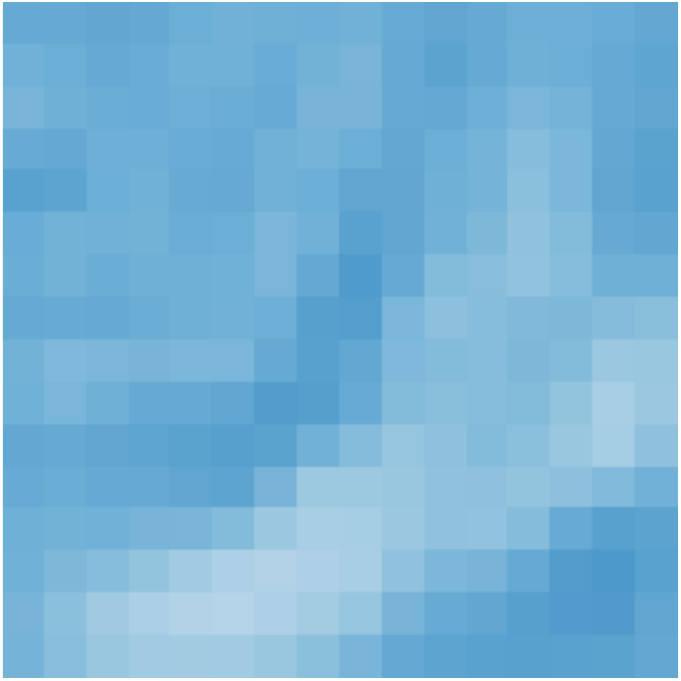}
	\vspace{-0.2cm}
	\caption{2D wave approximation in time step 45 provided by differently trained DISTANA4 models. From left to right: ideal wave (target), noisy wave with \SI{0.25}{SNR} (to be filtered), DISTANA4 trained on \SI{0.25}{SNR} and initialized with 30 steps teacher forcing, DISTANA4 trained on \SI{100}{SNR} and initialized with 30 steps teacher forcing, DISTANA4 trained on \SI{100}{SNR} and initialized with 30 active tuning steps.}
	\label{experiments_and_results:fig:active_tuning_waves}
\end{figure}

\autoref{experiments_and_results:fig:active_tuning_activity} shows the wave activity at a single position in the two-dimensional wave grid and visualizes the filtering capabilities of the model in three scenarios: (a) Model1 TF, trained and evaluated on \SI{0.25}{SNR}; initialized with TF, (b) Model2 TF trained on \SI{100}{SNR} and evaluated on \SI{0.25}{SNR}; initialized with TF, and (c) Model2 AT, trained on \SI{100}{SNR} and evaluated on \SI{0.25}{SNR}; initialized with AT.
The same scenarios are visualized spatially in one particular time step in \autoref{experiments_and_results:fig:active_tuning_waves}.
While Model1 TF is accurate during TF, it quickly fades to zero activity in closed loop application; which is reasonable since the very high training noise (\SI{0.25}{SNR}) forced it to developed a strong low-pass filter.
Model2 TF already produces inaccurate predictions in the TF phase and subsequently is incapable of continuing the signal reasonably; the model cannot deal with the strong noise, which it has never encountered during training.
However, the  same model initialized with AT (Model2 AT) can be used to produce highly detailed and accurate predictions in closed loop application without fading to zero or oscillating chaotically.
Previously mentioned video material emphasizes these findings.

The tuning parameters $\eta$ and $H$ were chosen carefully for each model.
Two trends can be observed in \autoref{experiments_and_results:tab:active_tuning}.
First, history length $H$ requires to be longer when $\text{SNR}_\text{train} < \text{SNR}_\text{test}$.
Second, tuning rate $\eta$ correlates with the training noise, that is, small training noise (large SNR) requires small choices of $\eta$.

\subsubsection{Drawback of active tuning}
Although AT yields impressive results when used to induce reasonable dynamics into a model, it requires explicit hand crafting of the tuning rate $\eta$, history length $H$, and number of optimization cycles (left constant at 30 in this work).
Additionally, AT comes with a massively increased computational cost when compared to TF, since instead of just forwarding an input through the network (TF), AT applies a local optimization procedure in every time step, which---depending on the chosen history length and optimization cycles---can slow down the activity induction process significantly.

%

\section{Discussion}

A spatio-temporal graph neural network (STGNN), DISTANA, designed to model and predict spatio-temporal processes---such as two-dimensional circular wave dynamics---has been compared to two state-of-the-art neural network structures: convolutional LSTMs (ConvLSTMs) and temporal convolution networks (TCNs).
The results show that DISTANA yields more accurate predictions and is mostly more robust against noise.
Furthermore, DISTANA was applied with active tuning (AT), which was used to successfully induce a stable state into the model and to replace the conventional teacher forcing (TF) procedure.

When trained on different signal to noise ratios (SNRs), the three architecture types differed in both train and test accuracy depending on the noise level.
As reported in \autoref{experiments_and_results:subsec:training_on_noisy_data}, ConvLSTM reached best results on extremely noisy data, while with decreasing noise, DISTANA was superior.
This behavior might likely come from the very basic architectural differences between the two models: while ConvLSTM has a spatial focus, DISTANA has a temporal focus.
More specifically, ConvLSTM first aggregates spatial input via the convolution operation and considers the temporal dimension solely on this aggregation.
DISTANA, on the other hand, first accesses the very local data, that is one pixel, while spatial aggregation is done via lateral temporal information exchange between Prediction Kernels (PKs).
This difference in processing order can explain ConvLSTM's slight advantage in highly noisy conditions (where information can be aggregated spatially), while DISTANA has the potential to very accurately approximate a specific dynamical process, once it is sufficiently interpretable.
Our results do not confirm recent findings \cite{kalchbrenner2016neural,dauphin2017language} which report TCNs as superior to recurrent neural networks in temporal information processing.
In our experiments, TCN never reached top performance.
Video material clearly shows the limits to which extent any explored architecture (even ConvLSTM) can factually model highly noisy data when explicitly trained on the particular noise level.

DISTANA was significantly less affected by overfitting (see \autoref{experiments_and_results:fig:mse_and_dtw_accuracy_comparison}).
While the test accuracy of ConvLSTM and TCN decreased on small training noise conditions (high SNR), DISTANA did not follow this trend, which can be explained by DISTANA's generalization abilities.
Apparently, DISTANA approximated the actual causal process with much smaller divergence in the training data vicinity, enabling it to properly switch from TF to closed loop application even in low-noise and noise-free training conditions.

We also applied DISTANA to the moving MNIST dataset \cite{srivastava2015unsupervised}.
DISTANA was not capable of identifying global characters and failed whenever a symbol touched the border and hence bounced off into the other direction.
Due to DISTANA's local connection scheme, it is not able to model the abrupt changes in motion direction of pixels that are far away from the border.
This issue will be addressed by extending the PK's neighborhood (e.g via skip connections) in future research.

Another essential finding of this work is that AT extends the applicability of any recurrent neural network model over the horizon of the training statistics.
The traditionally applied TF procedure fails here because a neural network model used with TF for activity induction generally cannot handle direct input that has a smaller SNR compared to the SNR it was trained on.
AT can bridge this gap by tuning the model states invariant to noise levels.
The necessary selection of convenient tuning parameters $\eta$ and $H$ and the additional computational overhead may be negligible in the light of the rather dramatic performance gains.

Overall, the long prediction stability over hundreds of closed loop steps produced by DISTANA is in line with findings from other works on GNNs, such as \cite{battaglia2016interaction}, who used a GNN to model the behavior of physical entities moving through space over thousands of time steps.

%

\section{Conclusion}

In conclusion, our evaluations show that DISTANA---a spatio-temporal graph neural network (STGNN)---outperforms ConvLSTM and TCN to large extents and in various respects. 
Our findings thus suggest that GNNs should be further applied to modeling spatio-temporal processes, which promises increased generalization, denoising, and deep, closed-loop future prediction abilities.  
Moreover, our results imply that active tuning may replace teacher forcing for the initialization of the latent activities in generative, recurrent neural network architectures, enabling models to be applied to larger noise than they were originally trained on. 
Future research needs to focus on evaluating this potential using real-world data such as soil, traffic, brain-imaging or social network data, while first results of ongoing research demonstrate DISTANA's applicability to global weather data.
%
%
%
\bibliographystyle{splncs04}
\bibliography{2020-denoisingDISTANA}
\end{document}